\documentclass[runningheads]{llncs}
\usepackage{siunitx}
\usepackage{eccv}

\usepackage{eccvabbrv}

\usepackage{xcolor}

\usepackage{graphicx}
\usepackage{booktabs}
\usepackage{multirow}
\usepackage{tabularx}
\usepackage{array}
\usepackage{comment}
\usepackage{caption}

\usepackage[accsupp]{axessibility}  

\usepackage{hyperref}

\usepackage{orcidlink}

\begin{document}

\title{DiffGI: Differentiable Geometry Images for High-Fidelity Thin-Shell 3D Generation}

\titlerunning{DiffGI: Differentiable Geometry Images}

\author{Eungjune Shim\orcidlink{0009-0006-5974-1542} \and
Hansol Lee\orcidlink{0009-0005-5708-6073} \and
Eunjung Ju\orcidlink{0000-0002-4549-9277}}


\authorrunning{E. Shim et al.}

\institute{CLO Virtual Fashion Inc., South Korea \\
\email{e.june.shim@gmail.com, \{river, kate\}@clo3d.com}}
\maketitle

\begin{abstract}
 Existing 3D generative models predominantly rely on implicit volumetric representations, which inherently enforce watertight topology and struggle to faithfully represent thin-shell and non-manifold geometries such as garments. While geometry image-based approaches offer a surface-centric alternative, existing methods typically rely on discrete binary occupancy maps whose resolution-dependent boundary encoding causes staircase artifacts and information loss upon downsampling, while surface reconstruction remains a non-differentiable post-processing step disconnected from the learning pipeline.

 To address this, we propose Differentiable Geometry Image (DiffGI)\footnote{Project page: \url{https://ejshim.github.io/diffgi/}}, an end-to-end 3D-to-2D mapping framework that seamlessly integrates surface representation and geometric optimization. DiffGI replaces conventional binary maps with a continuous 2D Truncated Signed Distance Function (TSDF), which encodes boundary position at subpixel precision within a fixed grid resolution, effectively eliminating resolution-dependent staircase artifacts even under aggressive downsampling. Building on this continuous field, we introduce a differentiable Marching Squares algorithm based on analytical linear interpolation, allowing gradients from 3D surface losses to propagate seamlessly back to the 2D latent space.

 Leveraging this differentiable pipeline, we train a DiffGI-VAE augmented with a geometry-aware normal rendering loss to compress complex 3D surfaces into an ultra-compact $32 \times 32$ latent space. Finally, we instantiate a transformer-based latent diffusion model on top of this space for conditional 3D generation, showing that the proposed representation readily supports efficient generative modeling. Extensive experiments on garment and object datasets demonstrate that our method achieves superior reconstruction fidelity and boundary precision compared to prior geometry-image and voxel-based approaches, while requiring significantly fewer computational resources.

\end{abstract}
\newpage
\section{Introduction}
\label{sec:intro}
Recent advances in combining large-scale data with deep learning have driven rapid progress in 3D generation. Neural implicit field representations such as SDFs, volumetric occupancy, and NeRFs learn continuous 3D spatial functions, enabling differentiable volume rendering and 3D distillation from 2D diffusion models.

However, these methods generally assume watertight surfaces, posing fundamental limitations in representing thin-shell and open-boundary structures critical in practical applications. When generating digital garments or furniture frames using implicit fields, networks often fail to maintain thin surfaces, resulting in artificial thickness or front-back blending. Moreover, meshes obtained via Marching Cubes lack UV coordinates, making them difficult to use in industrial pipelines requiring physics simulation and material editing.

Geometry image representations---which parameterize 3D surfaces onto 2D grids (specifically, multi-chart geometry images~\cite{sander2003multichart} that partition a surface into multiple UV charts)---offer an alternative, treating coordinates and attributes as image-like tensors and enabling direct use of 2D generative models. Omages~\cite{yan2025omages}, GIMDiffusion~\cite{elizarov2025gimdiffusion}, and GarmageNet~\cite{li2025garmagenet} have demonstrated this approach for UV-friendly 3D generation. However, existing methods rely on binary occupancy maps that are resolution-dependent and induce staircase artifacts, often requiring high-resolution grids (e.g., $768 \times 768$~\cite{elizarov2025gimdiffusion}) to partially compensate. Moreover, post-processing steps such as boundary snapping are non-differentiable, creating a structural disconnect between learning and mesh reconstruction.

Meanwhile, Differentiable Marching Cubes, DMTet, and FlexiCubes have made iso-surface extraction differentiable for implicit representations, but operate on 3D grids with cubic memory scaling and without natural UV parameterization. Our goal is to bring the advantages of differentiable iso-surface extraction to the geometry image family on 2D UV grids, achieving lighter computation and UV-friendly meshes simultaneously.

\begin{figure}[ht]
    \centering
    \includegraphics[width=0.8\textwidth]{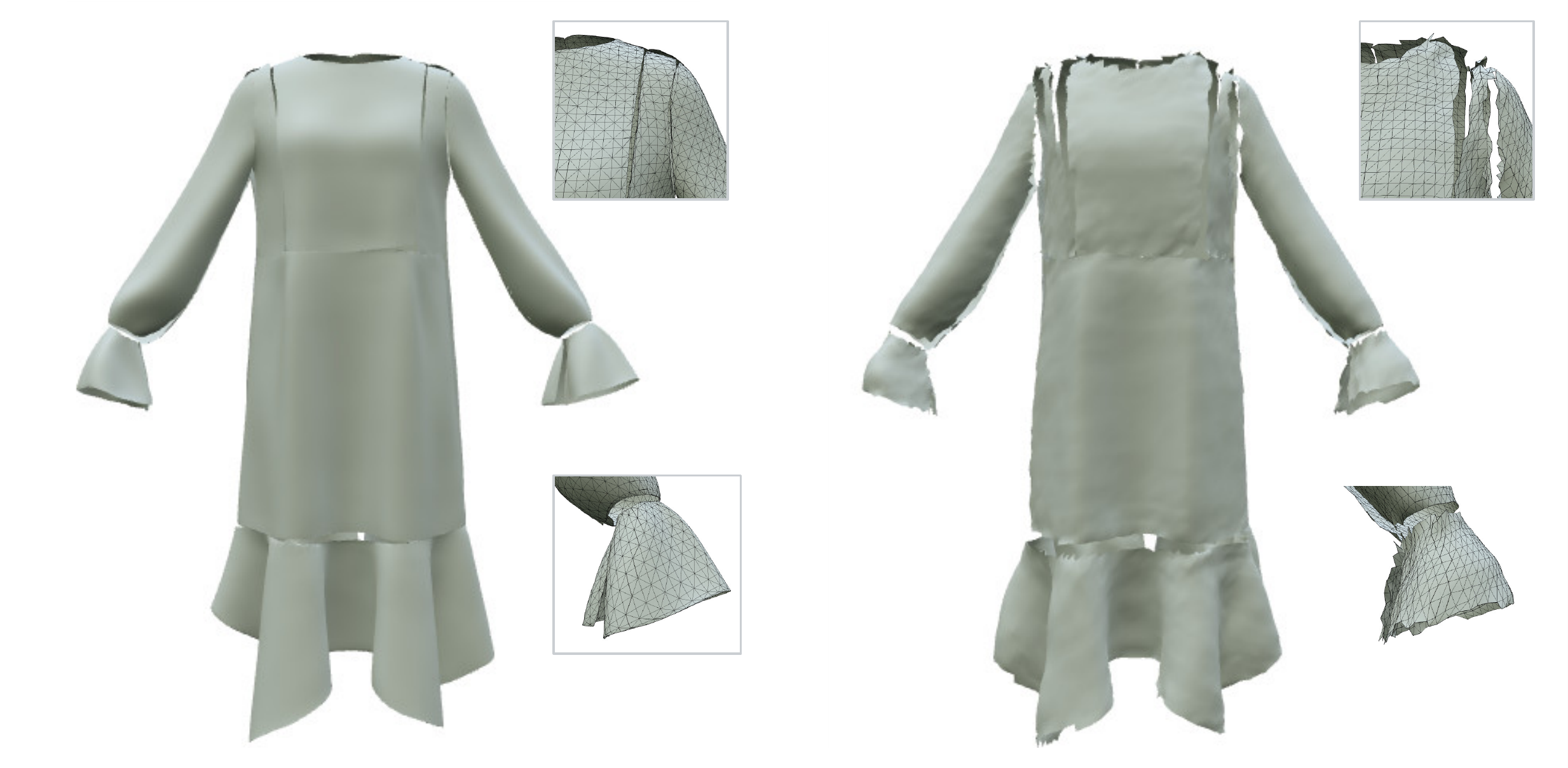}
    \caption{VAE reconstruction results with our TSDF-based DiffGI representation and normal rendering loss (left) versus an occupancy-based geometry image without normal loss (right), showing noticeably sharper boundaries and better preservation of thin-shell structures.}
    \label{fig:reconablation}
\end{figure}

\begin{figure}[ht]
    \centering
    \includegraphics[width=\textwidth]{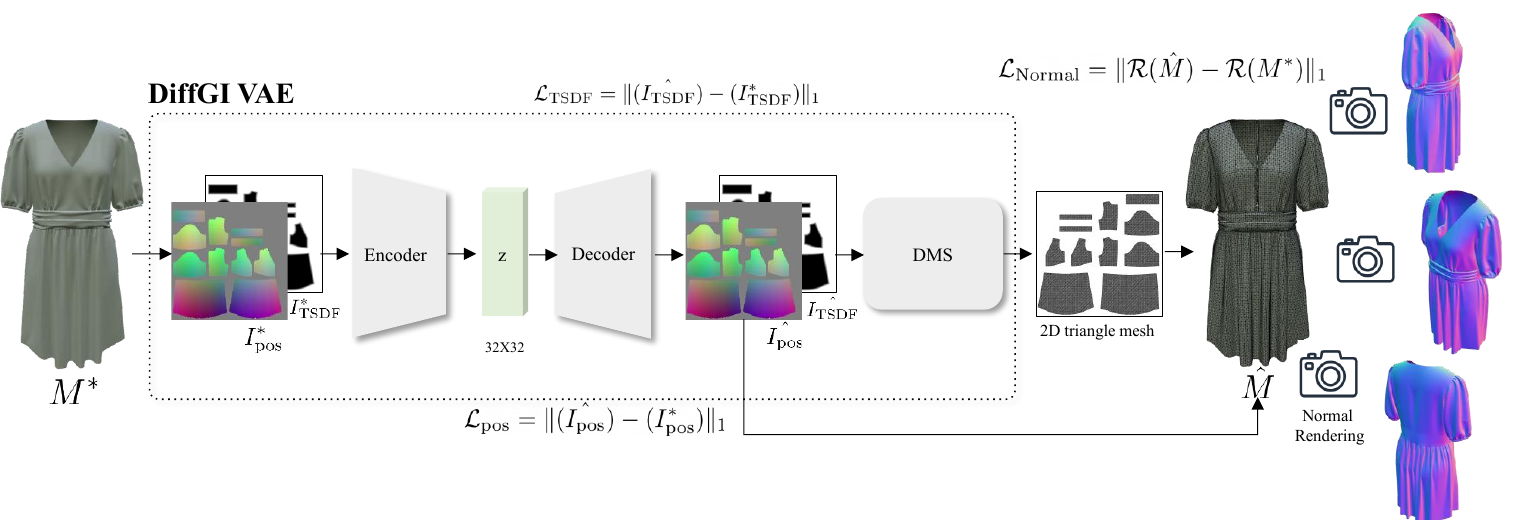}
    \caption{Overview of the proposed DiffGI framework. The input 3D mesh is first mapped to a 2D TSDF-based DiffGI representation, which is then encoded by a DiffGI-VAE into a compact latent space. The decoder reconstructs the TSDF geometry image, from which a 3D triangle mesh is recovered via Differentiable Marching Squares. Pixel-space losses on the TSDF and position maps, together with a geometry-aware normal rendering loss, are jointly applied to enable end-to-end optimization of high-fidelity 3D surfaces.}
    \label{fig:overview}
\end{figure}

We propose Differentiable Geometry Image (DiffGI), which replaces binary occupancy with a continuous 2D TSDF and introduces Differentiable Marching Squares (DMS) for fully differentiable mesh reconstruction. This allows gradients from 3D surface losses to propagate seamlessly to the 2D latent space. We build a VAE with geometry-aware normal rendering loss and a transformer-based latent diffusion model, demonstrating efficient generation of thin non-manifold meshes in an ultra-compact $32\times 32$ latent space (Figure~\ref{fig:overview}). As illustrated in Figure~\ref{fig:reconablation}, these design choices yield noticeably sharper boundaries and better preservation of thin-shell structures compared to occupancy-based methods.

Our contributions are:
\begin{enumerate}
    \item \textbf{Subpixel 2D Boundary Representation:} A geometry-image representation that replaces the binary occupancy maps of prior work with a continuous 2D TSDF. While the advantage of signed distance fields over occupancy grids is well established for volumetric 3D representations, we realize it in the 2D multi-chart geometry-image setting, where the continuous field encodes boundary position at subpixel precision within a fixed grid resolution and removes the resolution dependency and staircase artifacts of occupancy maps.
    \item \textbf{Fully Differentiable Mesh Reconstruction:} A tensor-based Differentiable Marching Squares algorithm that makes 3D reconstruction amenable to backpropagation, bridging the gap between 2D optimization and 3D mesh recovery.
    \item \textbf{Ultra-Compact Latent Space for Efficient Generation:} The TSDF representation and geometry-aware normal rendering loss enable compression of complex non-manifold surfaces into a $32{\times}32{\times}4$ latent space, supporting real-time 3D generation even on consumer-grade hardware.
\end{enumerate}

\section{Related Work}
\label{sec:related_work}

\subsection{Implicit and Explicit Representations for 3D Generation}
Recent 3D generation has primarily built on implicit field representations---signed/unsigned distance fields, occupancy fields, and radiance fields---which provide continuous geometry and integrate naturally with diffusion- or rendering-based supervision~\cite{park2019deepsdf, chen20223psdf, wang2022hsdf, cheng2023sdfusion, chou2023diffusionsdf, zheng2023locally, long2023neuraludf, meng2023neat, zhou2024udiff, fainstein2024dudf, chen2023singlestagenerf, gu2022stylenerf}, enabling a wave of large-scale 3D generation and reconstruction systems~\cite{hong2024lrm, wang2024crm, xu2024instantmesh, yang2024hunyuan3d, zhao2025hunyuan3d2, xiang2025structured, li2025triposg, boss2025sf3d}.
UDF-based methods relax the closed-surface bias of SDFs~\cite{long2023neuraludf, meng2023neat, zhou2024udiff, fainstein2024dudf}, but remain field-based and still require dedicated mesh extraction and post-processing~\cite{hou2023dcudf, xie2025ldm, wang2024genudc}, limiting their applicability to thin garment panels, open boundaries, and downstream tasks such as simulation and material authoring.
Explicit surface representations preserve mesh structure and UV parameterization, making them more compatible with graphics pipelines~\cite{gu2002geometryimages, yang2024dreammesh, yu2024texgen}, yet their irregular discrete topology makes end-to-end modeling challenging.
Our work retains a structured 2D surface representation while enabling differentiable, boundary-aware surface recovery.

\subsection{Geometry-Image-Based Surface Representations}
Geometry images map irregular 3D surfaces onto regular UV-aligned 2D grids, enabling surface coordinates and attributes to be treated as image-like tensors~\cite{gu2002geometryimages, sander2003multichart, carr2006rectmcgim, sinha2016deepgim}.
Recent methods such as GIMDiffusion~\cite{elizarov2025gimdiffusion}, Omages~\cite{yan2025omages}, and GarmageNet~\cite{li2025garmagenet} show that this representation pairs well with 2D generative backbones for UV-friendly 3D generation.
However, these pipelines still rely on binary occupancy or mask channels~\cite{elizarov2025gimdiffusion, yan2025omages, li2025garmagenet}, making boundary localization resolution-dependent and often requiring heuristics such as boundary snapping or chart masking.
DiffGI retains the geometry-image prior but replaces binary support with a continuous 2D TSDF on a single $256{\times}256$ tensor, enabling continuous boundary localization instead of hard mask-based recovery.

\subsection{Differentiable Iso-Surface Extraction}
Differentiable iso-surface extraction makes field-to-surface conversion part of the optimization loop.
Starting from Marching Cubes~\cite{lorensen1987marchingcubes}, numerous methods have pursued this goal, notably DMTet~\cite{shen2021dmtet} and FlexiCubes~\cite{shen2023flexicubes}, along with others~\cite{liao2018deepmc, chen2021nmc, chen2022ndc, hou2023dcudf, liu2024gshell, son2024dmesh, son2025dmeshpp, binninger2025tetweave, stippel2025marchingneurons}.
These methods consistently demonstrate that surface-level supervision is more effective when extraction participates in optimization.
However, most operate on 3D voxel or tetrahedral grids, where memory scales cubically and UV-aligned surfaces are not directly available~\cite{liao2018deepmc, chen2021nmc, chen2022ndc, shen2021dmtet, shen2023flexicubes, hou2023dcudf, binninger2025tetweave}.
DiffGI brings this idea to geometry images: we introduce Differentiable Marching Squares (DMS) over a 2D TSDF in UV space, combining end-to-end surface optimization with the efficiency and UV compatibility of structured 2D representations.
\section{Method}
\label{sec:method}
\subsection{Continuous TSDF Geometry Image Representation}

Prior geometry-image-based methods combine a position map encoding 3D coordinates with a binary occupancy map indicating the valid region. However, binary occupancy maps encode boundaries as hard 0/1 transitions, making boundary localization inherently resolution-dependent: upon downsampling to practical training resolutions, fine boundary details are destroyed and staircase artifacts emerge.

To address this, the DiffGI representation replaces the binary occupancy map with a \textbf{Truncated Signed Distance Function (TSDF)} defined as a continuous function on the 2D plane. Because the TSDF encodes the distance to the nearest boundary rather than a binary label, it preserves subpixel boundary information even at low resolution, enabling high-fidelity reconstruction from aggressively compressed grids. The Mesh-to-DiffGI conversion is performed as an offline
preprocessing step before training, and proceeds as follows:

\begin{enumerate}
    \item \textbf{UV Packing \& Global Scaling:} UV patches are arranged on the 2D plane using an AABB-based packing algorithm with a uniform global scale and inter-patch padding to prevent bleeding, iteratively optimized to maximize fill ratio.
    \item \textbf{Spatial Sampling \& Position Map:} The mesh surface is sampled on a $1024 \times 1024$ regular grid via barycentric interpolation, producing a 3-channel position map.
    \item \textbf{Dilation:} Edge pixel values are propagated outward into the undefined background via dilation, preventing boundary corruption during downsampling.
    \item \textbf{2D TSDF Transform:} For every pixel on the $1024 \times 1024$ UV grid, including those in the empty background outside the charts, we compute the signed Euclidean distance in pixels to the nearest chart contour (positive inside a chart, negative outside), clamped at a truncation distance of 15 pixels.
    \item \textbf{Bilinear Downsampling:} The $1024 \times 1024$ tensor is downsampled to $256 \times 256 \times 4$. Unlike binary maps, the continuous TSDF ensures no information destruction or staircase artifacts during downsampling.
\end{enumerate}

\subsection{Differentiable Marching Squares (DMS)}

When recovering a 3D mesh from the 2D DiffGI tensor generated within a deep learning architecture (DiffGI-to-Mesh), the reconstruction process must be differentiable with respect to the network weights in order to backpropagate geometric losses from 3D space to the 2D model. However, the traditional Marching Squares algorithm determines topology through a discrete lookup table, blocking gradients at this stage and making learning within the deep learning computation graph infeasible.

To address this, we propose the Differentiable Marching Squares (DMS) module, which formulates vertex positions as continuous functions of the underlying TSDF field. On the 2D TSDF grid, when the corner values $\phi_A$ and $\phi_B$ of two adjacent pixels have opposite signs ($\phi_A \cdot \phi_B < 0$), the Intermediate Value Theorem guarantees that a surface boundary (zero-crossing) exists between them. The relative position $x \in [0, 1]$ at which the boundary point lies is derived through linear interpolation. To ensure numerical stability, we employ the following interpolation formula with a regularization constant $\epsilon$:
\begin{equation}
x = \frac{\phi_A}{\phi_A - \phi_B + \epsilon \cdot \text{sgn}(\phi_A - \phi_B)}
\end{equation}
Here, $\epsilon$ (e.g., $10^{-5}$) serves to prevent gradient explosion in regions where the two pixel values are nearly equal ($\phi_A \approx \phi_B$), which would otherwise cause the denominator to approach zero.
Since $\text{sgn}(\cdot)$ is locally constant and thus treated as a
non-differentiable constant during backpropagation, gradients flow
solely through the continuous ratio
$\phi_A / (\phi_A - \phi_B)$, preserving a smooth optimization
landscape. The subpixel vertex 2D UV coordinates extracted through this formula are then transformed into final 3D vertex coordinates $V = (X, Y, Z)$ via bilinear grid sampling on the 3-channel position map tensor.

The key advantage of this algorithm lies in its optimizability during the backward pass. The geometric error (e.g., normal loss) $\mathcal{L}$ computed at the reconstructed 3D vertices $V$ is smoothly propagated back to the 2D TSDF tensor and position map tensor in the form of $\frac{\partial \mathcal{L}}{\partial V}$ via the chain rule. Among the 16 topological configurations of Marching Squares, Case~6 (BR+TL) and Case~9 (BL+TR)---where two pairs of diagonal corners have opposite signs---correspond to topologically ambiguous saddle point cases. In these cases, both interpretations of connecting or separating the two regions are mathematically valid. In this work, we deterministically adopt the convention of always treating the two patches as separate independent regions. While the topological configuration choice itself is discrete, the coordinates of each boundary vertex within the chosen topology remain continuous functions of the TSDF values according to Eq.~(1). Therefore, the backpropagation path is fully preserved even in ambiguous cases. The module is implemented in a vectorized fashion using PyTorch tensor operations, guaranteeing significantly faster computation at $O(N^2)$ complexity compared to the $O(N^3)$ complexity of existing 3D volumetric extraction methods (DMC, DMTet, etc.).

\subsection{Geometry-Aware Latent Compression (DiffGI-VAE)}

Directly generating high-quality 3D surfaces in the high-dimensional tensor space of $256 \times 256 \times 4$ resolution is limited in terms of memory and training efficiency. We introduce the DiffGI-VAE framework to compress this data into a much smaller and more compact space. To accelerate convergence, we initialize the DiffGI-VAE with pretrained VAE weights from Stable Diffusion 1.5~\cite{rombach2022ldm}, which has learned general-purpose spatial compression priors from large-scale natural images. The channel count of the first convolutional layer is expanded to accommodate the 4-channel input scheme (Position 3 + TSDF 1), and zero-initialization is applied to the newly added weights to prevent catastrophic forgetting.

To faithfully preserve 3D geometric shapes while aggressively compressing the latent space dimensions to $32 \times 32 \times 4$, we design the total loss function $\mathcal{L}_{total}$ as follows:
\begin{equation}
    \mathcal{L}_{total} = \mathcal{L}_{Pos} + \lambda_{TSDF} \mathcal{L}_{TSDF} + \lambda_{Normal} \mathcal{L}_{Normal} + \lambda_{KL} \mathcal{L}_{KL}
\end{equation}
Here, $\mathcal{L}_{Pos}$ and $\mathcal{L}_{TSDF}$ are L1 losses that enforce pixel-level reconstruction accuracy. However, pixel-level losses alone are insufficient to adequately constrain surface curvature and orientation information, leaving the preservation of high-frequency geometric features---such as fine fabric wrinkles and sharp edges---unguaranteed. To address this, we introduce a \textbf{geometry-aware normal rendering loss ($\mathcal{L}_{Normal}$)}. The mesh differentiably reconstructed through the DMS module is rendered into a normal map image using the differentiable rasterizer nvdiffrast~\cite{laine2020diffrast}, and the L1 error against the normal map identically rendered from the ground truth mesh is directly minimized as follows:
\begin{equation}
    \mathcal{L}_{Normal} = \lVert \mathcal{R}(\hat{M}) - \mathcal{R}(M^{*}) \rVert_1
\end{equation}
where $\mathcal{R}(\cdot)$ denotes the nvdiffrast rendering function, $\hat{M}$ is the reconstructed mesh, and $M^{*}$ is the ground truth mesh. This encourages high-frequency geometric features to be compressed into and reconstructed from the latent space without distortion. The advantages of combining the TSDF-based representation with the geometry-aware normal rendering loss over occupancy-based representations, in terms of boundary sharpness and preservation of thin non-manifold structures, are quantitatively examined in the ablation study in Section~\ref{sec:Experiments}.

\subsection{Latent Diffusion on DiffGI Latent Space}

To verify that the proposed representation naturally supports high-quality conditional 3D generation on the $32 \times 32$ latent space constructed by DiffGI-VAE, we train a transformer-based latent diffusion model.

Unlike natural photographs, geometry image tensors require capturing global topological connectivity rather than relying on spatial inductive biases of pixel positions. We therefore depart from the U-Net architecture and adopt the Diffusion Transformer (DiT)~\cite{peebles2023dit} as the core backbone, which allows unrestricted global attention among input tokens. We also employ a flow-matching~\cite{lipman2023flowmatching} formulation-based scheduler, which has been shown to achieve high-quality generation with fewer inference steps. To support diverse downstream applications, we construct three conditional generation models:

\begin{enumerate}
    \item \textbf{Label-Conditioned Generation:} Generates global shapes conditioned on input class labels (e.g., furniture, lamp). A DiT-B/2 architecture is used to comprehensively learn shape characteristics.
    \item \textbf{Image-Conditioned Generation:} Infers unseen back surfaces from a single front-view image to generate a complete 3D mesh. Semantic embedding vectors extracted by the DINOv2-Large~\cite{oquab2024dinov2} vision foundation model are injected into DiT-L/2 via a cross-attention mechanism to perform complex shape prediction.
    \item \textbf{Occupancy-Conditioned Generation:} Generates physically plausible 3D garment draping and wrinkles conditioned on 2D sewing patterns or silhouette information. Since the input condition already strongly constrains the spatial structure in this task, preserving local spatial features is relatively more important than global attention. Accordingly, instead of a global attention-based DiT, a lightweight UNet-Tiny architecture with skip connections is selectively adopted to ensure computational efficiency.
\end{enumerate}

To mitigate positional bias from fixed UV layouts, we apply geometric
data augmentation via random $90^\circ$-multiple rotations and
re-packing at the UV chart level during training. These transformations
alter only the 2D layout while preserving the geometric signals within
each chart, effectively expanding the training distribution and
improving generalization (visual examples and quantitative ablation
in the supplementary material).

\section{Experiments}
\label{sec:Experiments}
\subsection{Experimental Setup and Evaluation Metrics}

\paragraph{Datasets.}
We evaluate on three benchmarks: \textbf{ABO}~\cite{collins2022abo} (7,900 commercial 3D furniture scans with complex topologies and open structures), \textbf{GarmageSet}~\cite{li2025garmagenet} (14,801 physics-simulation-ready 3D garments with strong non-manifold characteristics), and \textbf{WARDROBE}~\cite{wardrobe_vision_dataset_2025} (${\sim}$300 real clothing images used for qualitative evaluation of zero-shot image-to-3D generalization).

\paragraph{Evaluation Metrics.}
For per-sample geometric evaluation, we randomly sample 100,000 points from both the generated and ground truth mesh surfaces. We report \textbf{Chamfer Distance (CD)} for macroscopic shape accuracy, \textbf{Normal Consistency (NC)}, computed as one minus the average L1 distance between normal maps rendered from four fixed viewpoints (so that higher is better), and \textbf{F1-score} at a distance threshold of 0.01. To assess boundary quality, we use \textbf{Boundary Chamfer Distance (BCD)}, which computes CD only on open-boundary points, and \textbf{Hausdorff Distance (HD, $d_H$)} for worst-case local distortion. For distribution-level evaluation, we measure \textbf{P-FID} and \textbf{P-KID} using pretrained PointNet++~\cite{qi2017pointnet++} features, along with \textbf{EMD} and \textbf{JSD} between generated and real point cloud distributions. Table~\ref{tab:vae_eval} summarizes the reconstruction and encoding fidelity of all methods.

\begin{table}[ht]
\centering
\captionsetup{font=scriptsize}
\caption{Quantitative comparison of 3D reconstruction and encoding
fidelity on the ABO and GarmageSet datasets. We report
CD, EMD, JSD, and NC for Omages, GarmageNet, and our DiffGI-VAE. All baselines use the same Omages-style tessellation (Tess.); for GarmageNet we additionally report its native official extraction (Official).
For GarmageNet, $N$ denotes the number of garment panels, each
represented as a fixed 72-dimensional vector.}
\label{tab:vae_eval}
\setlength{\tabcolsep}{3pt}
\resizebox{\textwidth}{!}{%
\begin{tabular}{l c cccc cccc}
\toprule
    \multirow{2}{*}{\textbf{Method}} 
        & \multirow{2}{*}{
            \begin{tabular}[c]{@{}c@{}}Representation\\ Size\end{tabular}} 
            & \multicolumn{4}{c}{ABO Dataset} 
            & \multicolumn{4}{c}{GarmageSet} \\ 
    \cmidrule(lr){3-6} \cmidrule(l){7-10} 
        & 
        & CD ($\times 10^{-3}$) $\downarrow$ & EMD $\downarrow$ & JSD ($\times 10^{-3}$) $\downarrow$ & NC $\uparrow$
        & CD ($\times 10^{-3}$) $\downarrow$ & EMD $\downarrow$ & JSD ($\times 10^{-3}$) $\downarrow$ & NC $\uparrow$ \\ 
    \midrule
        omages64 & $64{\times}64{\times}4$
            & 0.89 & 0.25 & 0.92 &  \textbf{0.89} 
            & 1.31 & 0.17 & 1.79 & 0.95  \\
        GarmageNet (Tess.) & $N$ $\times$ 72
            & - & - & - & - 
            & 2.19 & 0.21 & 32.61 &  0.88 \\ 
            GarmageNet (Official) & $N$ $\times$ 72
            & - & - & - & -
            & 1.89 & 0.17 & 5.51 & 0.90
            \\
    \midrule 
        \textbf{ours} & $32{\times}32{\times}4$
        & \textbf{0.83} & \textbf{0.23} & \textbf{0.89} & 0.83 
        & \textbf{0.46} & \textbf{0.16} & \textbf{1.24} & \textbf{0.96} \\ 
    \bottomrule
\end{tabular}
}
\end{table}

\begin{figure}[ht] 
    \centering
    \includegraphics[width=\textwidth]{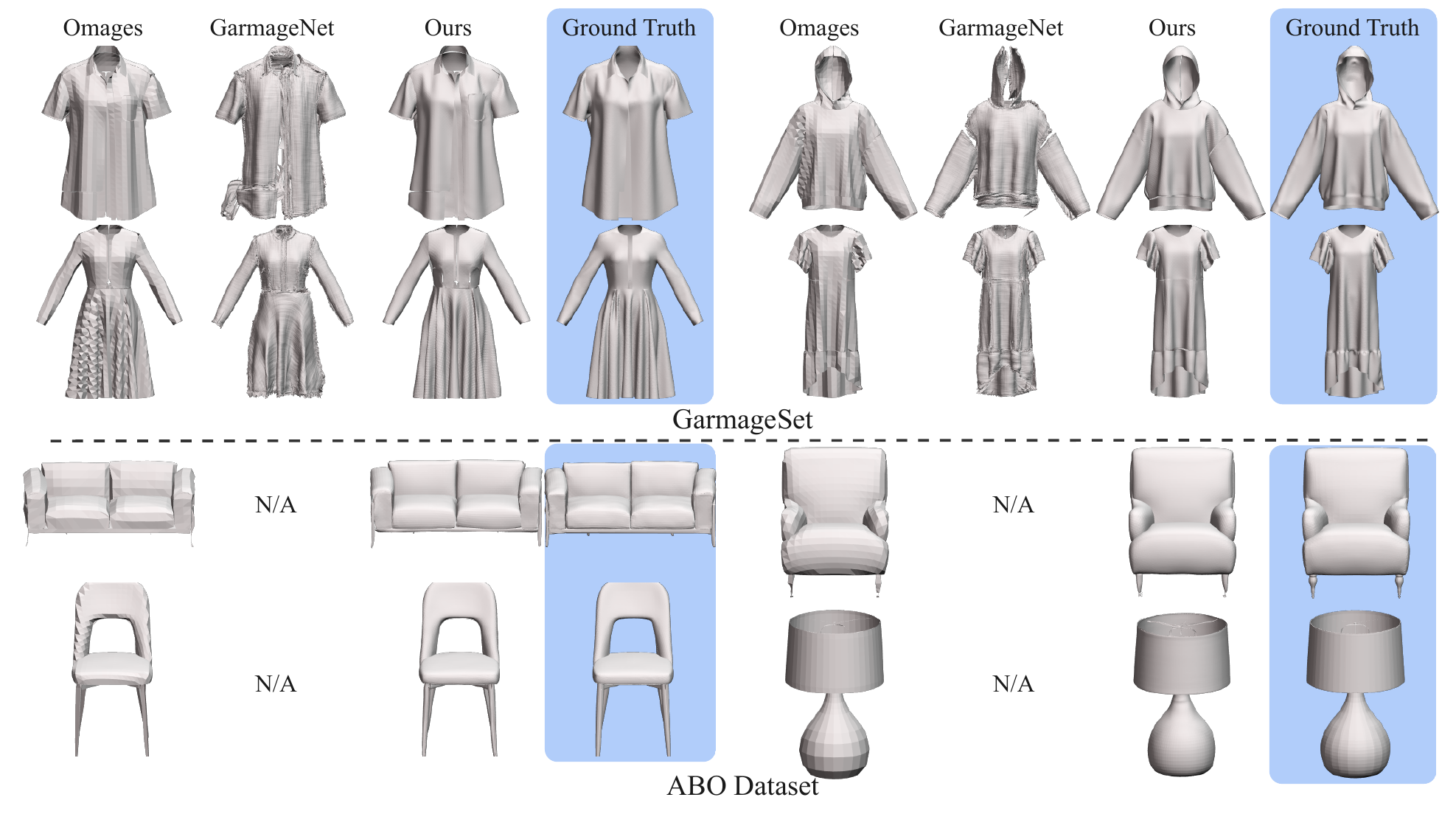} 
    \caption{Qualitative comparison of VAE reconstructions on the ABO and GarmageSet datasets, illustrating that our DiffGI-VAE yields sharper boundaries and better preservation of thin-shell details than Omages and GarmageNet.}
    \label{fig:Garmageset} 
\end{figure}

\subsection{Reconstruction Fidelity (DiffGI-VAE)}

We compare DiffGI-VAE against Omages and GarmageNet on the ABO and GarmageSet datasets. Omages directly converts meshes to geometry images without VAE compression; GarmageNet uses per-panel geometry images with a fixed panel count, making it inapplicable to ABO (entries excluded). To evaluate each representation with its intended pipeline, each method uses its native mesh extractor: Differentiable Marching Squares for DiffGI and the Omages-style tessellation for Omages. Since GarmageNet's native extraction is a slow multi-stage non-learning procedure (${\sim}$5.1\,s per mesh), we report it under the same Omages-style tessellation (Table~\ref{tab:vae_eval}, Tess.) for the main comparison and additionally under its native official extraction (Official). DiffGI remains superior under both GarmageNet settings.

As shown in Table~\ref{tab:vae_eval} and Figure~\ref{fig:Garmageset},
DiffGI-VAE consistently outperforms baselines in CD, EMD, and JSD on both
datasets. Notably, our method compresses a $256{\times}256$ geometry image
into a $32{\times}32{\times}4$ VAE latent, whereas Omages directly operates
on a $64{\times}64{\times}4$ geometry image without VAE compression;
despite this aggressive compression, DiffGI-VAE achieves superior
reconstruction fidelity. On ABO, NC is slightly lower than Omages,
which we attribute to the information loss inherent in $8{\times}$
spatial compression, particularly on flat furniture surfaces where
Omages' uncompressed representation retains normal information more
directly. On GarmageSet, however, our method achieves the highest NC,
confirming state-of-the-art reconstruction quality for thin non-manifold
garment structures. This demonstrates that the continuous TSDF
representation with fully differentiable optimization effectively
preserves curvature and fine details even under substantial latent
compression.

\begin{table}[ht]
\centering
\captionsetup{font=scriptsize}
\caption{Ablation study of representation (Occ. vs. TSDF) and normal rendering loss (NL) for our DiffGI-VAE on GarmageSet. TSDF-based variants outperform occupancy-based ones across all metrics, and enabling NL yields the highest NC with the lowest CD, EMD, and JSD.}
\label{tab:ablation_vae}

\scriptsize
\setlength{\tabcolsep}{1.6pt}          
\renewcommand{\arraystretch}{0.82}     
\begin{tabular}{@{}l c cccc@{}}
\toprule
Rep. & $\mathcal{L}_{\text{Normal}}$ & CD ($\times 10^{-3}$) $\downarrow$ & EMD $\downarrow$ & JSD ($\times 10^{-3}$) $\downarrow$ & NC $\uparrow$ \\
\midrule
Occ  & $\times$      & 1.503 & 0.171 & 4.539 & 0.906 \\
Occ  & \checkmark    & 1.313 & 0.166 & 4.257 & 0.947 \\
TSDF & $\times$      & 0.595 & 0.165 & 2.169 & 0.921 \\
TSDF & \textbf{\checkmark} & \textbf{0.461} & \textbf{0.160} & \textbf{1.244} & \textbf{0.961} \\
\bottomrule
\end{tabular}
\end{table}

\begin{figure}[ht]
    \centering
    \includegraphics[width=\textwidth]{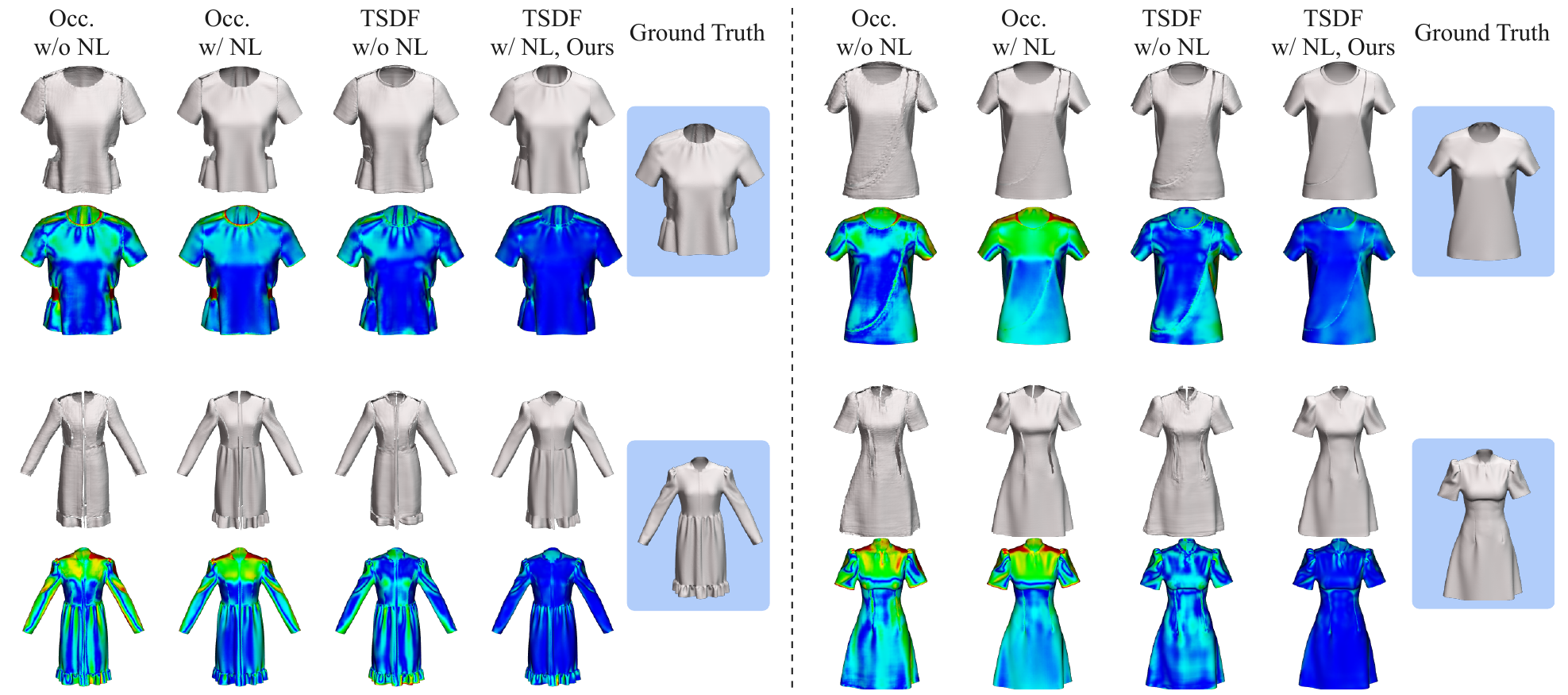} 
    \caption{Qualitative ablation study of our VAE on the GarmageSet dataset. The visual comparisons demonstrate how different representations (Occ. vs. TSDF) and the normal rendering loss affect the 3D reconstruction quality.}
    \label{fig:VAE_ablation} 
\end{figure}

\subsection{Ablation Study: Differentiable Mesh Reconstruction and Normal Loss}

To verify that the performance gains of the DiffGI framework result from the cumulative effect of each design component, we conduct an ablation study that systematically decomposes the contributions of the representation (Occ.\ vs.\ TSDF) and the geometry-aware normal rendering loss ($\mathcal{L}_{Normal}$). Since occupancy maps do not admit differentiable mesh extraction by default, we reconstruct the occupancy variants with a differentiable formulation of the Omages-style tessellation, so that all four variants receive the same normal rendering supervision and the comparison isolates the representation rather than the extraction step.

The results in Table~\ref{tab:ablation_vae} demonstrate that each of the two design components independently yields significant performance gains.

\textbf{Effect of representation (Occ.\ vs.\ TSDF).}
When only the representation is switched from Occ.\ to TSDF without $\mathcal{L}_{Normal}$, CD decreases by more than half from $1.503 \times 10^{-3}$ to $0.595 \times 10^{-3}$, and JSD also improves substantially from $4.539 \times 10^{-3}$ to $2.169 \times 10^{-3}$. This quantitatively confirms that the continuous TSDF encodes geometric information in boundary regions far more effectively than binary occupancy maps.

\textbf{Effect of geometry-aware normal rendering loss ($\mathcal{L}_{Normal}$).}
Enabling $\mathcal{L}_{Normal}$ on the same representation consistently improves all metrics. In particular, under the TSDF-based setting, NC increases from 0.921 to 0.961, with CD and JSD further improving as well. Notably, the NC of the Occ.\ + $\mathcal{L}_{Normal}$ combination (0.947) exceeds the NC of TSDF without $\mathcal{L}_{Normal}$ (0.921), suggesting that $\mathcal{L}_{Normal}$ enforces normal orientation information strongly enough to partially compensate for the limitations of the representation. However, in terms of CD, EMD, and JSD, TSDF without $\mathcal{L}_{Normal}$ still outperforms Occ.\ + $\mathcal{L}_{Normal}$, indicating that the contribution of the representation itself is more fundamental for overall shape reconstruction.

As shown in Figure~\ref{fig:VAE_ablation}, the Occ.\ without $\mathcal{L}_{Normal}$ setting---trained with pixel-level losses only---exhibits prominent high-frequency noise and aliasing artifacts at mesh boundaries and surfaces. In the final configuration combining the TSDF representation with $\mathcal{L}_{Normal}$, these artifacts are substantially mitigated and boundaries are reconstructed with noticeably greater sharpness.

\textbf{Effect of VAE initialization.}
We additionally train the same DiffGI-VAE architecture from random initialization and observe nearly identical final reconstruction fidelity to the SD1.5-initialized model on GarmageSet (CD 0.47 vs.\ 0.46, NC 0.96 for both), confirming that the reconstruction gains stem from the DiffGI representation and differentiable reconstruction pipeline rather than the SD1.5 VAE initialization; detailed results and convergence analysis are provided in Sec.~2 of the supplementary material.

 \begin{table}[t]
\centering
\captionsetup{font=scriptsize}
\caption{Computational Efficiency \& Inference Performance. We compare peak memory usage and inference latency across different model scales, conditioning tasks, and hardware setups. Our DiffGI framework demonstrates exceptional efficiency, capable of running even on a consumer-grade CPU.}
\label{tab:efficiency}

\scriptsize
\setlength{\tabcolsep}{3.5pt}
\renewcommand{\arraystretch}{0.88}

\begin{tabular}{@{}llcc@{}}
\toprule
Method & Hardware & Peak VRAM (GB) $\downarrow$ & Time (sec) $\downarrow$ \\
\midrule
TRELLIS-image        & RTX A6000 Ada     & 16.28 & 4.52 \\
TRELLIS.2 512$^{2}$  & RTX A6000 Ada     & 2.65  & 12.35 \\
TRELLIS.2 1024$^{2}$ & RTX A6000 Ada     & 6.41  & 45.14 \\
GarmageNet           & RTX A6000 Ada     & 0.96  & 0.40 \\
Omages               & RTX A6000 Ada     & 2.49  & 52.0 \\
\midrule
Ours-Label           & RTX A6000 Ada     & 1.18  & 0.50 \\
Ours-Image           & RTX A6000 Ada     & 3.22  & 0.80 \\
Ours-Image           & RTX 4070 (12GB)   & 3.22  & 1.21 \\
Ours-Image           & MacBook M4 (CPU)  & --    & 8.52 \\
\bottomrule
\end{tabular}
\end{table}

\subsection{Computational Efficiency and Inference Performance}

Table~\ref{tab:efficiency} compares inference cost across methods using each method's official recommended settings. Volumetric models such as TRELLIS require up to 16.28\,GB VRAM, and Omages takes 52 seconds per sample. In contrast, DiffGI operates in a compact 2D latent space, completing image-conditioned generation in ${\sim}$1.2\,s on a consumer GPU (RTX 4070, 3.22\,GB VRAM) and ${\sim}$8.5\,s on CPU only (MacBook M4), demonstrating scalability from servers to edge devices.

\begin{table}[tb]
\captionsetup{font=scriptsize}
  \caption{Label-conditioned 3D generation performance on the ABO dataset, reported as P-FID and P-KID per category and averaged (Mean).}
  \label{tab:label_condition_eval}
  \centering

  \scriptsize
  \setlength{\tabcolsep}{3.0pt}          
  \renewcommand{\arraystretch}{0.88}     

  \begin{tabular}{@{}lcccccccccc@{}}     
    \toprule
      & \multicolumn{2}{c}{Chair}
      & \multicolumn{2}{c}{Lamp}
      & \multicolumn{2}{c}{Sofa}
      & \multicolumn{2}{c}{Table}
      & \multicolumn{2}{c}{Mean} \\
    \cmidrule(lr){2-3}\cmidrule(lr){4-5}\cmidrule(lr){6-7}\cmidrule(lr){8-9}\cmidrule(lr){10-11}
      & Omages & Ours
      & Omages & Ours
      & Omages & Ours
      & Omages & Ours
      & Omages & Ours \\
    \midrule
    P-FID
      & 22.30 & \textbf{11.15}
      & 52.48 & \textbf{26.55}
      & 15.81 & \textbf{10.49}
      & 33.99 & \textbf{31.45}
      & 31.14 & 19.91 \\
    P-KID
      & 0.08 & \textbf{0.03}
      & 0.26 & \textbf{0.07}
      & 0.06 & \textbf{0.04}
      & \textbf{0.10} & 0.11
      & 0.12 & 0.06 \\
    \bottomrule
  \end{tabular}
\end{table}

\begin{figure}[ht] 
    \centering
    \includegraphics[width=\textwidth]{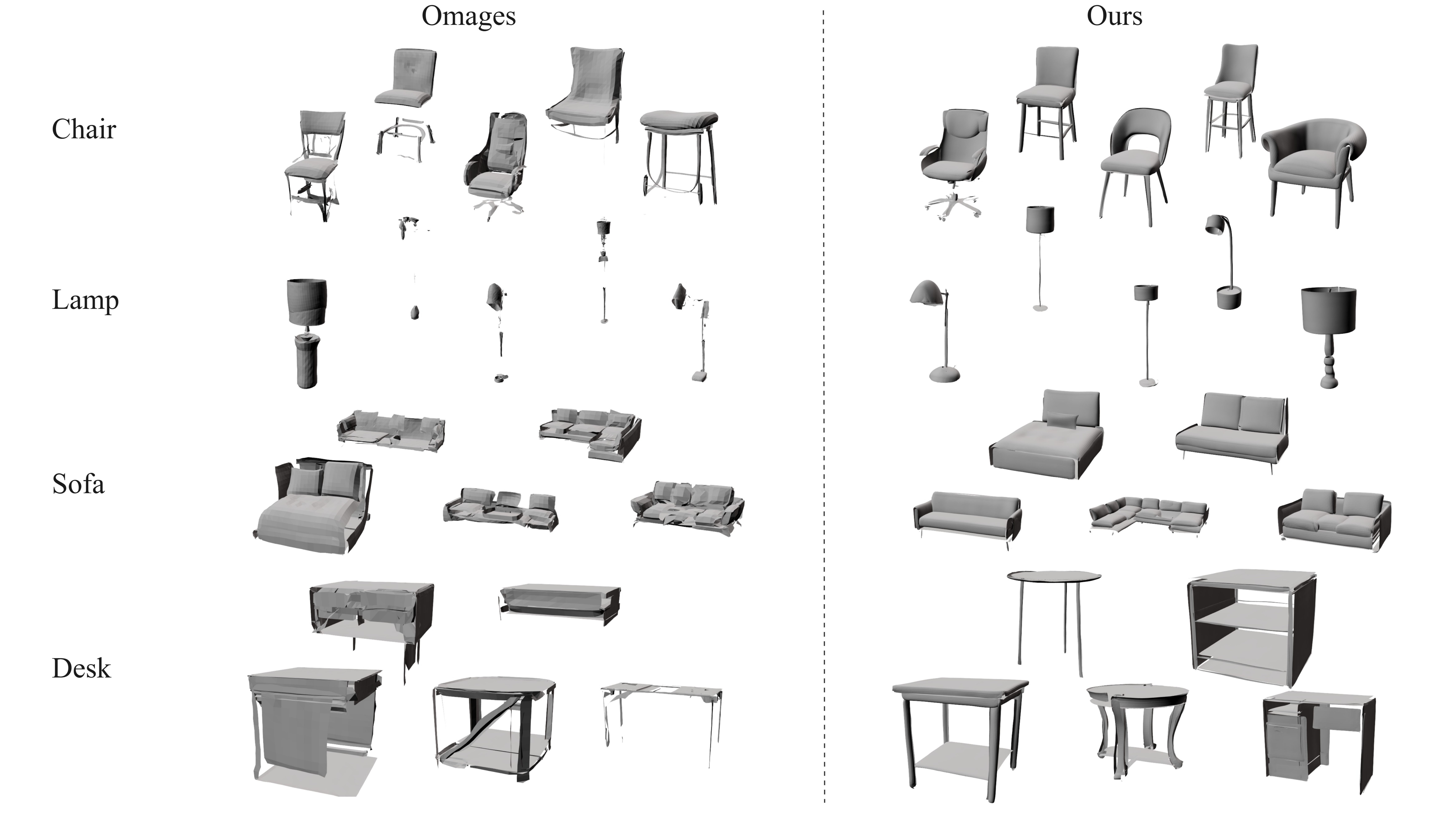} 
    \caption{Qualitative label-conditioned generation results on the ABO dataset. We compare Omages and our DiffGI-based diffusion model, showing improved reconstruction of thin frames and open-boundary structures, with fewer staircase artifacts.}
    \label{fig:ABO}
\end{figure}

\subsection{Downstream Generative Tasks: Open-surface 3D Generation}
\subsubsection{Label-Conditioned Generation}
We generate 512 3D samples per class on ABO and measure P-FID and P-KID. As shown in Table~\ref{tab:label_condition_eval} and Figure~\ref{fig:ABO}, the proposed model consistently improves P-FID across all categories, with the largest reductions in Chair (50\%) and Lamp (49\%). For Lamp, the absolute P-FID remains elevated due to limited training samples, but P-KID (0.07 vs.\ Omages 0.26) confirms robust generation quality. Qualitatively, Omages produces broken thin frames due to binary masks, while DiffGI generates smooth, continuous structures via subpixel boundary estimation.

\begin{table}[t]
\centering
\captionsetup{font=scriptsize}
\caption{Quantitative comparison of image-to-3D generation on
GarmageSet. TRELLIS and TRELLIS.2 are general-purpose foundation
models trained on large-scale multi-category 3D data, whereas
GarmageNet and our DiffGI are trained specifically on GarmageSet.
We report this comparison to highlight the practical advantages of
a surface-centric representation for thin-shell garment geometry,
rather than to claim universal superiority.}
\label{tab:image_conditioned}
\resizebox{\textwidth}{!}{
\begin{tabular}{lccccc}
\toprule
    \textbf{Method}
    & \textbf{\#Vert.\,$\downarrow$}
    & \textbf{CD ($\times 10^{-2}$) $\downarrow$}
    & \textbf{F1 $\uparrow$}
    & \textbf{HD ($\times 10^{-2}$) $\downarrow$}
    & \textbf{BCD ($\times 10^{-2}$) $\downarrow$} \\
\midrule
TRELLIS                & 109K & $3.44 \pm 6.97$ & $0.28 \pm 0.14$ & $15.38 \pm 27.57$ & N/A \\
TRELLIS.2 512$^{2}$    & 380K & $11.01 \pm 7.64$ & $0.27 \pm 0.14$ & $69.70 \pm 50.00$ & $12.44 \pm 8.56$ \\
GarmageNet             & 526K & $4.31 \pm 3.03$ & $0.20 \pm 0.12$ & $23.76 \pm 11.12$ & $5.64 \pm 2.94$ \\
\textbf{Ours (DiffGI)} & \textbf{23K} & $\textbf{1.35} \pm \textbf{0.47}$ & $\textbf{0.48} \pm \textbf{0.12}$ & $\textbf{8.42} \pm \textbf{3.25}$ & $\textbf{2.91} \pm \textbf{0.83}$ \\
\bottomrule
\end{tabular}
}
\medskip
\end{table}

\begin{figure}[ht]
    \centering
    \includegraphics[width=\textwidth]{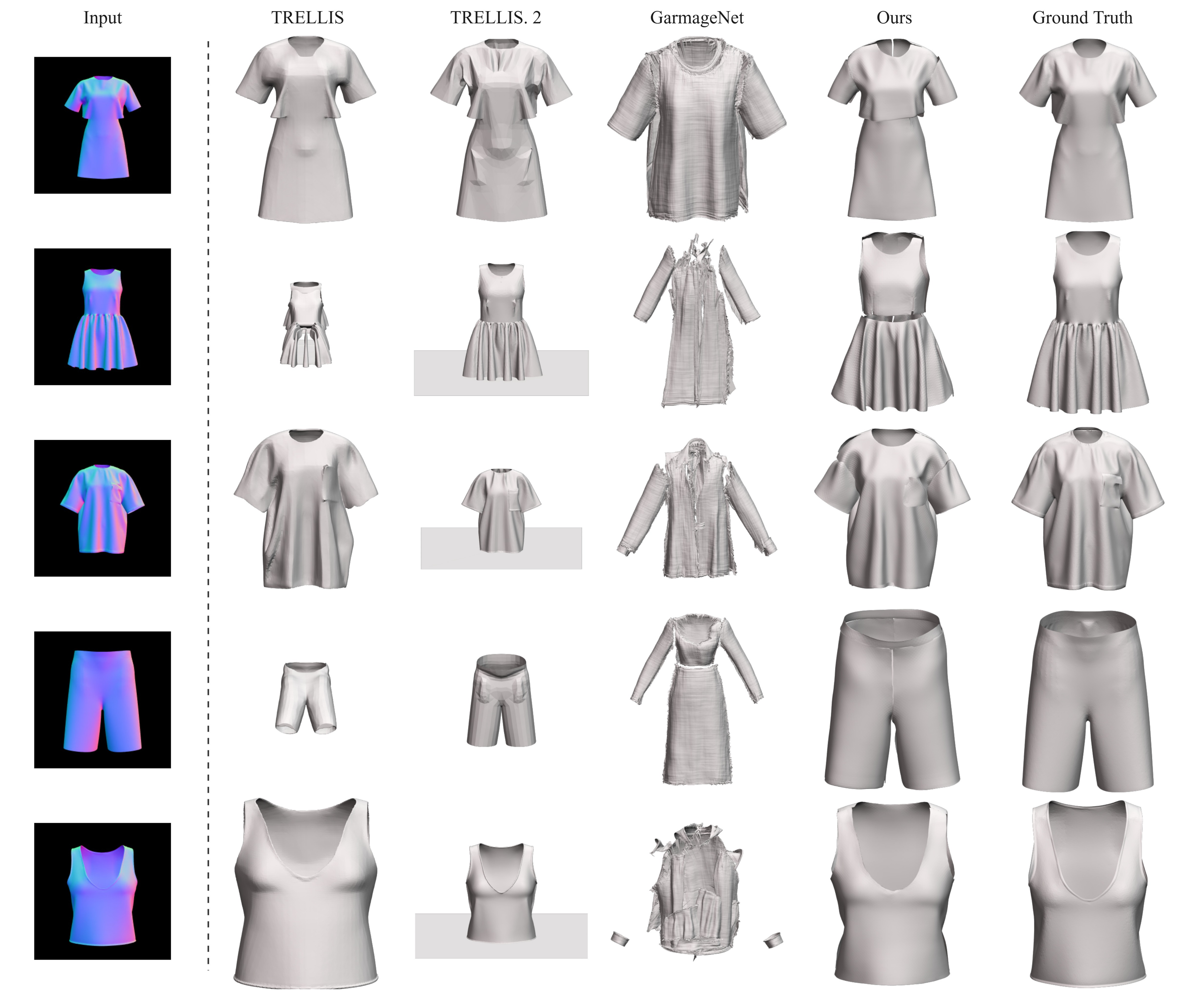} 
    \caption{Qualitative single-view image-to-3D comparison on GarmageSet. From left to right, each row shows the input front-view rendered normal map, the results of TRELLIS, TRELLIS.2, and GarmageNet, our DiffGI generation, and the ground-truth mesh. DiffGI preserves thin-shell details and produces cleaner open boundaries with far fewer vertices than the watertight foundation models (TRELLIS, TRELLIS.2) and the occupancy-based GarmageNet.}
    \label{fig:image_condition_diffusion} 
\end{figure}

\subsubsection{Image-Conditioned Generation: Comparison with Foundation Models}

For the \emph{single-view image-to-3D} task, we compare DiffGI against foundation-based models (TRELLIS~\cite{xiang2025structured}, TRELLIS.2~\cite{xiang2025trellis2}) and GarmageNet~\cite{li2025garmagenet} on GarmageSet. Our goal is not to claim universal superiority, but to examine the practical advantages of a surface-centric representation for thin-shell and open-boundary garment geometry.

As shown in Table~\ref{tab:image_conditioned}, DiffGI achieves the best CD ($1.35 \times 10^{-2}$), F1 (0.48), $d_H$ ($8.42 \times 10^{-2}$), and BCD ($2.91 \times 10^{-2}$) with only 23K vertices on average. The low BCD in particular confirms that the TSDF-based representation and differentiable reconstruction pipeline stably preserve thin fabric boundaries. Qualitatively (Figure~\ref{fig:image_condition_diffusion}), TRELLIS tends to produce excessive thickness on thin-shell garments due to its watertight assumption, while GarmageNet exhibits residual boundary aliasing from binary occupancy maps. DiffGI reconstructs cleaner boundaries with more compact meshes.

Although our image-conditioned model is trained on front-view rendered normal maps, at inference it also generalizes to real RGB front-view photographs, as shown by the zero-shot results on the WARDROBE dataset in the supplementary material, demonstrating the extensibility of our representation to diverse conditioning inputs.

\begin{figure}[ht]
    \centering
    \includegraphics[width=\textwidth]{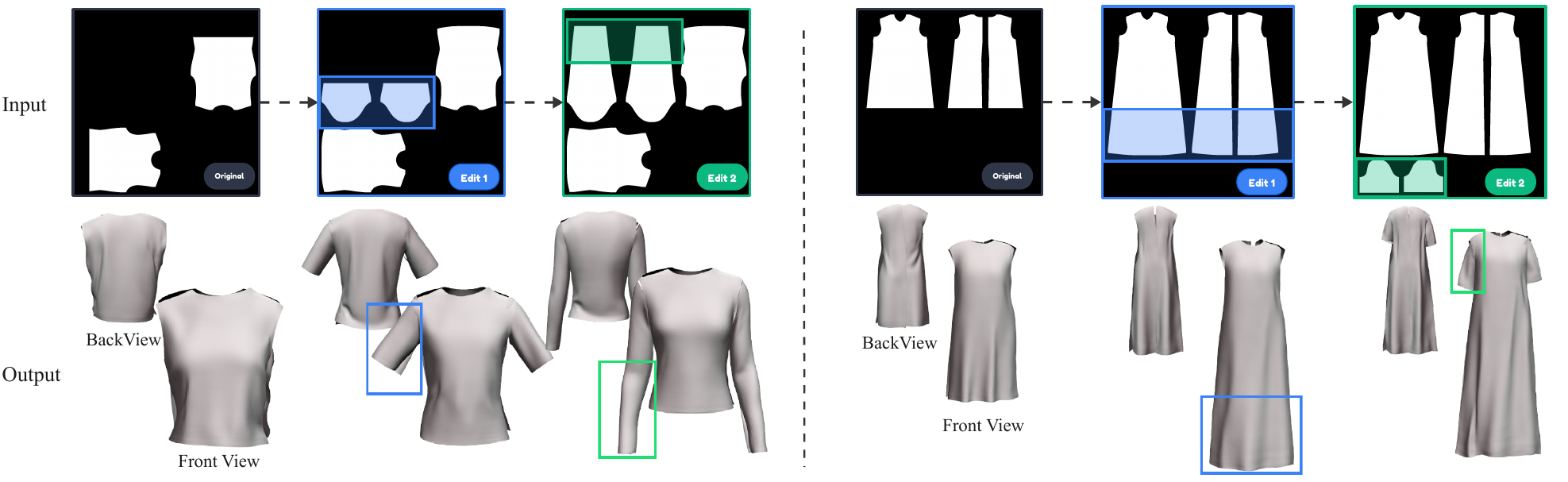} 
    \caption{Occupancy-conditioned garment generation and local edit propagation. Given a 2D occupancy map as input, modifying only the sleeve region in 2D leads to consistent updates of the corresponding 3D geometry, demonstrating that local pattern edits are faithfully reflected in the generated garment mesh.}
    \label{fig:draper}
\end{figure}

\subsubsection{Occupancy-Conditioned Generation}
We generate 3D garments conditioned solely on 2D sewing pattern silhouettes. Despite using a lightweight UNet-Tiny decoder, visually natural draping and wrinkles are stably produced (Figure~\ref{fig:draper}). Locally modifying only the sleeve region in the 2D occupancy map yields consistent updates to the corresponding 3D geometry while preserving the remaining patterns, demonstrating that 2D pattern-level edits directly propagate to 3D mesh variations.
\section{Conclusion, Limitations, and Future Work}
\label{sec:conclusion}

We proposed Differentiable Geometry Image (DiffGI), a fully differentiable framework that replaces binary occupancy maps with a continuous 2D TSDF, enabling subpixel-precise boundary reconstruction on geometry images. Our Differentiable Marching Squares (DMS) module brings the mesh reconstruction step inside the learning graph, allowing 3D surface losses to backpropagate seamlessly to the 2D latent space. Built on this, DiffGI-VAE compresses complex non-manifold surfaces into a $32{\times}32{\times}4$ latent space with a geometry-aware normal rendering loss, consistently outperforming prior geometry-image methods on both garment and furniture benchmarks. A transformer-based latent diffusion model trained on this space achieves competitive or superior generation quality with significantly less compute, running in real time even on consumer-grade hardware.

\paragraph{Limitations.}
TSDF-based linear interpolation can introduce localized rounding on extreme sharp edges (e.g., mechanical parts), and the current framework does not generate RGB textures or PBR attributes alongside geometry. In addition, since each UV chart is reconstructed independently and the DMS module treats adjacent patches as separate regions, the recovered geometry can show visible seams where neighboring charts meet, such as between the upper and lower parts of the dress in Figure~\ref{fig:image_condition_diffusion}. These boundary discontinuities do not affect our reconstruction and generation metrics, but they pose additional challenges for downstream physical simulation. Enforcing cross-chart boundary consistency is left for future work.

\paragraph{Future work.}
We plan to (1) integrate feature-preserving iso-surfacing (e.g., dual contouring) into DMS for sharp-edge domains, (2) extend the latent space to jointly generate high-resolution textures and material maps, and (3) introduce a two-stage pipeline separating 2D pattern layout generation from 3D shape reconstruction for stronger controllability over complex garment designs.

\section*{Acknowledgements}
We thank Hyun Kang, Seungoh Han, Sihun Cha, Dong-sig Kang, and Gyoo-Chul Kang for valuable discussions and feedback throughout this work. We are also grateful to CLO Virtual Fashion for providing the research environment and resources that made this work possible.

\par\vfill\par

%
%
\bibliographystyle{splncs04}
\bibliography{main}

\clearpage
\setcounter{section}{0}
\setcounter{figure}{0}
\setcounter{table}{0}
\renewcommand{\thesection}{S\arabic{section}}
\renewcommand{\thefigure}{S\arabic{figure}}
\renewcommand{\thetable}{S\arabic{table}}

\begin{center}
  {\Large\bfseries DiffGI: Differentiable Geometry Images for
    High-Fidelity Thin-Shell 3D Generation \par}
  \vspace{0.5em}
  {\large --- Supplementary Material --- \par}
\end{center}
\vspace{1em}

\section{Resolution-Dependent Analysis: TSDF vs.\ Occupancy}
\label{sec:supp_resolution}

To isolate the intrinsic effect of the representation choice independent of
VAE compression, we compare TSDF- and Occupancy-based geometry images
across resolutions (64, 128, 256, 512, 1024) \emph{without} VAE compression.

\begin{figure}[tb]
  \centering
  \includegraphics[width=\linewidth]{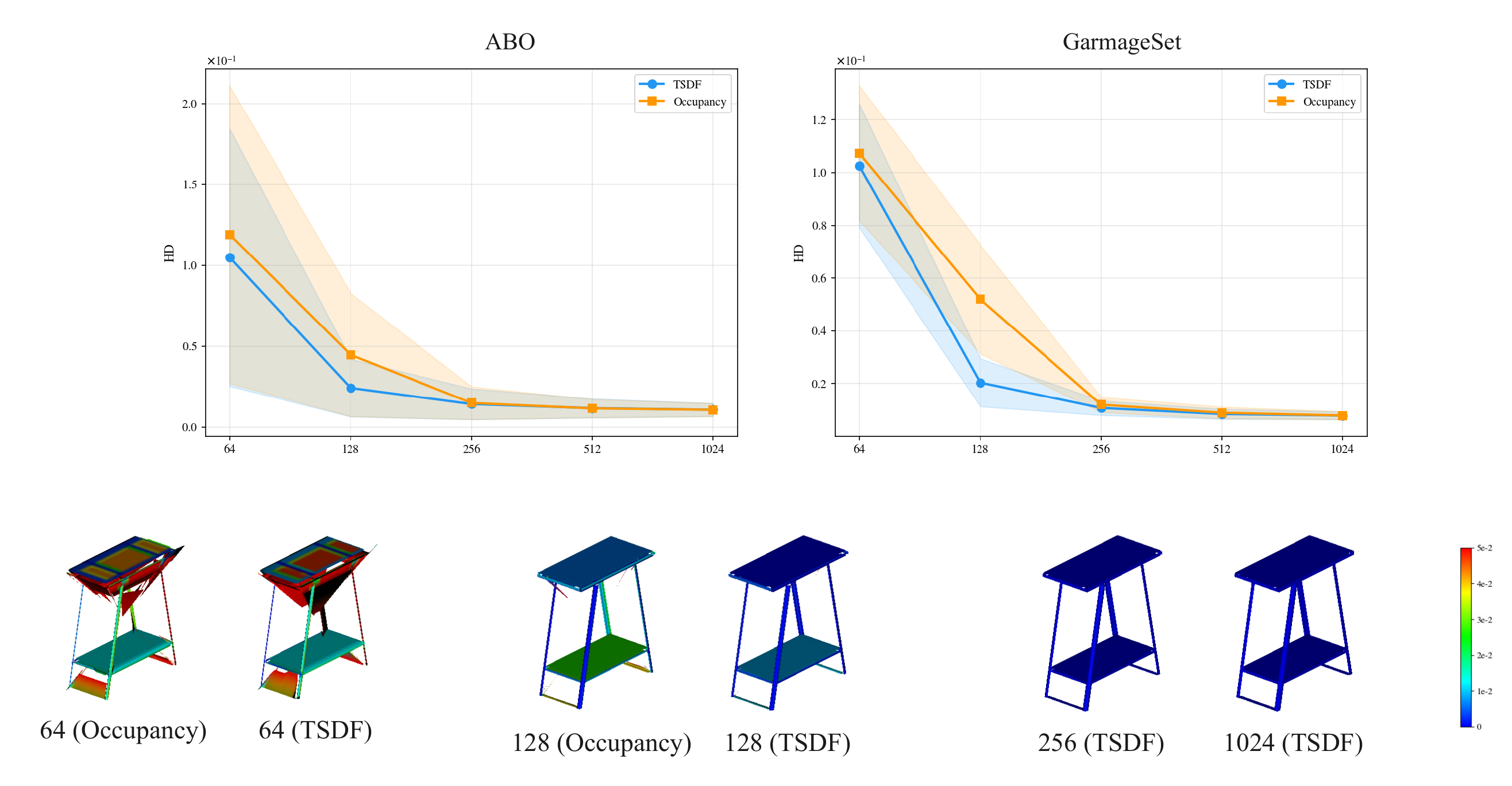}
  \caption{%
    Resolution-dependent reconstruction quality of TSDF vs.\ Occupancy
    without VAE compression.
    \textbf{Top:} Hausdorff Distance~(HD) vs.\ resolution on ABO~(left)
    and GarmageSet~(right); solid lines and shaded regions denote mean and
    standard deviation.
    \textbf{Bottom:} Per-vertex error maps for a representative ABO sample
    (colour scale: point-to-surface distance).
    TSDF achieves lower error at low resolutions (64, 128),
    with both representations converging beyond~256.%
  }
  \label{fig:resolution_hd}
\end{figure}

As shown in \cref{fig:resolution_hd}, TSDF consistently achieves lower
Hausdorff Distance~(HD) at low resolutions, with the gap most pronounced at
64 and 128, where Occupancy suffers from staircase artifacts along thin
structures and open boundaries. Beyond resolution~256, both representations
converge to near-identical error, confirming that 256 provides sufficient
reconstruction fidelity as the operating resolution for our pipeline.

\paragraph{Dataset-dependent behaviour.}
The TSDF advantage is more pronounced on GarmageSet---where garment meshes
exhibit dense wrinkles, curved hemlines, and highly irregular open
boundaries---than on ABO, whose furniture shapes are dominated by flat planes
and straight edges. This confirms that the benefit of continuous boundary
encoding grows with boundary complexity.

\paragraph{Implications for VAE compression.}
The continuous TSDF encodes boundary positions as real-valued distances spread
across neighbouring pixels, whereas binary occupancy concentrates all boundary
information in a single $0/1$ transition pixel. This redundancy makes TSDF
significantly more robust to VAE compression---consistent with the ablation in
the main paper (Table~2), where the modest raw difference at high resolution
is amplified into a substantial performance gap after compression.

\section{VAE Initialization: SD1.5 Pretraining vs.\ Training from Scratch}
\label{sec:supp_vae_init}

To examine how much of the reconstruction quality depends on the Stable
Diffusion~1.5 initialization, we additionally train the DiffGI-VAE from a
random initialization while keeping the architecture and all other settings
identical, and evaluate it on GarmageSet under DMS extraction
(\cref{tab:vae_init}).

\begin{figure}[tb]
  \centering
  \includegraphics[width=\linewidth]{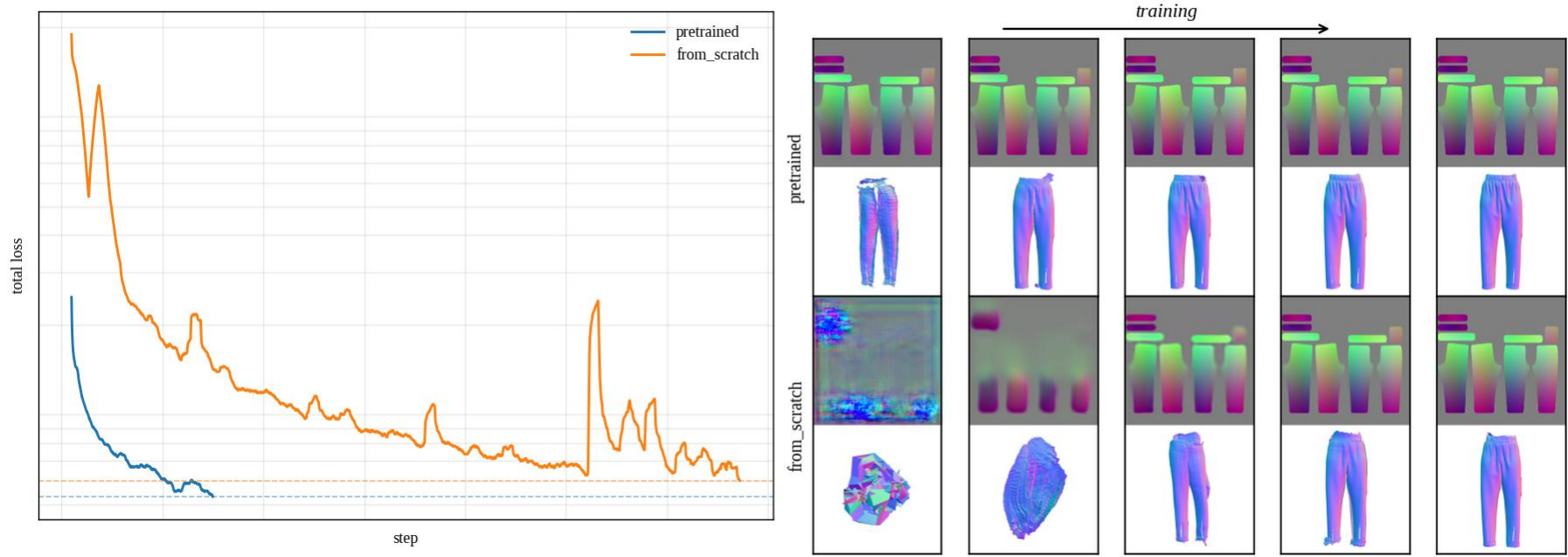}
  \caption{From-scratch vs.\ SD1.5-pretrained DiffGI-VAE.
    \textbf{Left:} total loss during early training; the SD1.5-initialized
    model (blue) reaches a low loss within far fewer steps, whereas the
    from-scratch model (orange) starts much higher and converges more slowly.
    \textbf{Right:} reconstructions over the course of training (top: geometry
    image, bottom: 3D mesh); the from-scratch model begins from noise but
    progressively recovers reconstructions essentially identical to the
    pretrained one. Both reach comparable final fidelity when fully trained
    (\cref{tab:vae_init}), so SD1.5 initialization mainly accelerates
    convergence rather than improving final quality.}
  \label{fig:vae_init}
\end{figure}

The from-scratch VAE reaches reconstruction fidelity comparable to the
SD1.5-initialized one (CD $0.47$ vs.\ $0.46$, identical NC of $0.96$),
indicating that the reconstruction gains stem from the TSDF representation
rather than from the pretrained initialization. The SD1.5 initialization
mainly accelerates convergence: as shown in \cref{fig:vae_init}, it reaches a
low loss within far fewer steps, while the from-scratch model starts from noise
and progressively recovers comparable reconstructions. Once both are trained to
convergence they attain essentially the same fidelity, which is why their final
metrics nearly coincide.

\begin{table}[tb]
  \caption{Effect of VAE initialization on GarmageSet reconstruction
    (DMS extraction, identical architecture). Training from scratch reaches
    fidelity comparable to SD1.5 initialization.}
  \label{tab:vae_init}
  \centering
  \small
  \begin{tabular}{@{}lcccc@{}}
    \toprule
    Initialization & CD ($\times 10^{-3}$) $\downarrow$ & EMD $\downarrow$
      & JSD ($\times 10^{-3}$) $\downarrow$ & NC $\uparrow$ \\
    \midrule
    From scratch    & 0.47 & 0.17 & 1.23 & 0.96 \\
    SD1.5 (default) & 0.46 & 0.16 & 1.24 & 0.96 \\
    \bottomrule
  \end{tabular}
\end{table}

\section{Impact of Geometric Data Augmentation and Model Scaling}
\label{sec:augmentation}

\begin{figure}[tb]
  \centering
  \includegraphics[width=\textwidth]{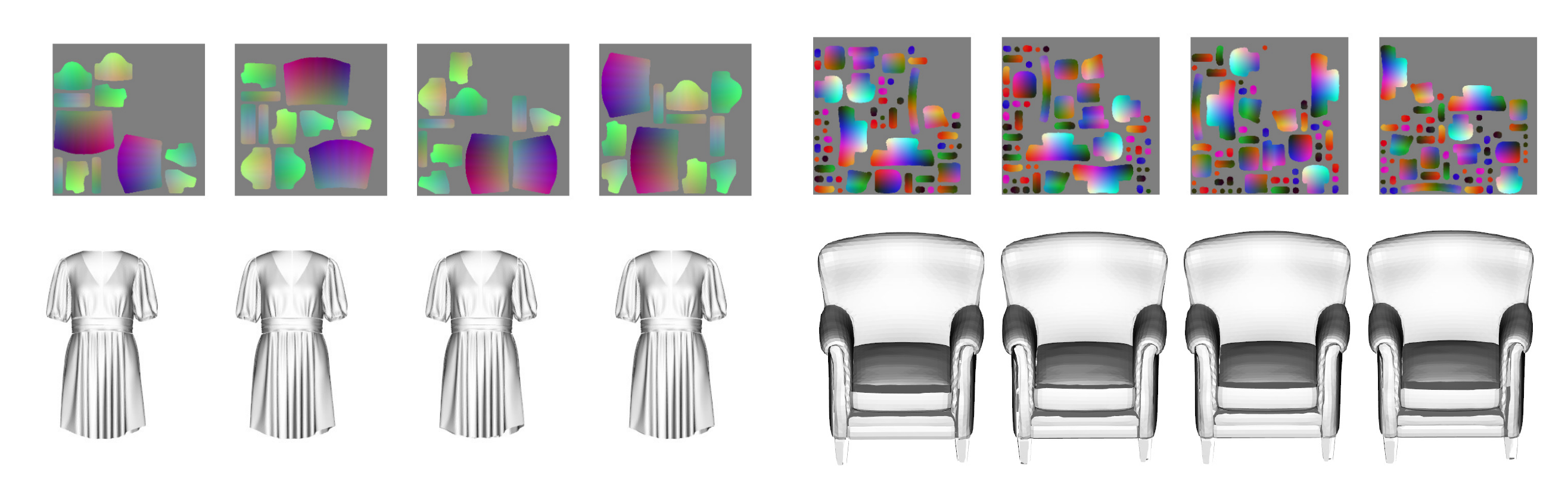}
  \caption{UV placement perturbation examples. Different UV chart layouts are
    generated from the same 3D mesh while preserving geometric signals,
    enabling large-scale data expansion without changing 3D semantics.}
  \label{fig:augmentation_method}
\end{figure}

We analyze the impact of geometric data augmentation (\eg, UV placement
perturbation) on training stability and model capacity, using the ABO dataset.
Visual examples are shown in \cref{fig:augmentation_method}.

\begin{figure}[tb]
  \centering
  \includegraphics[width=\textwidth]{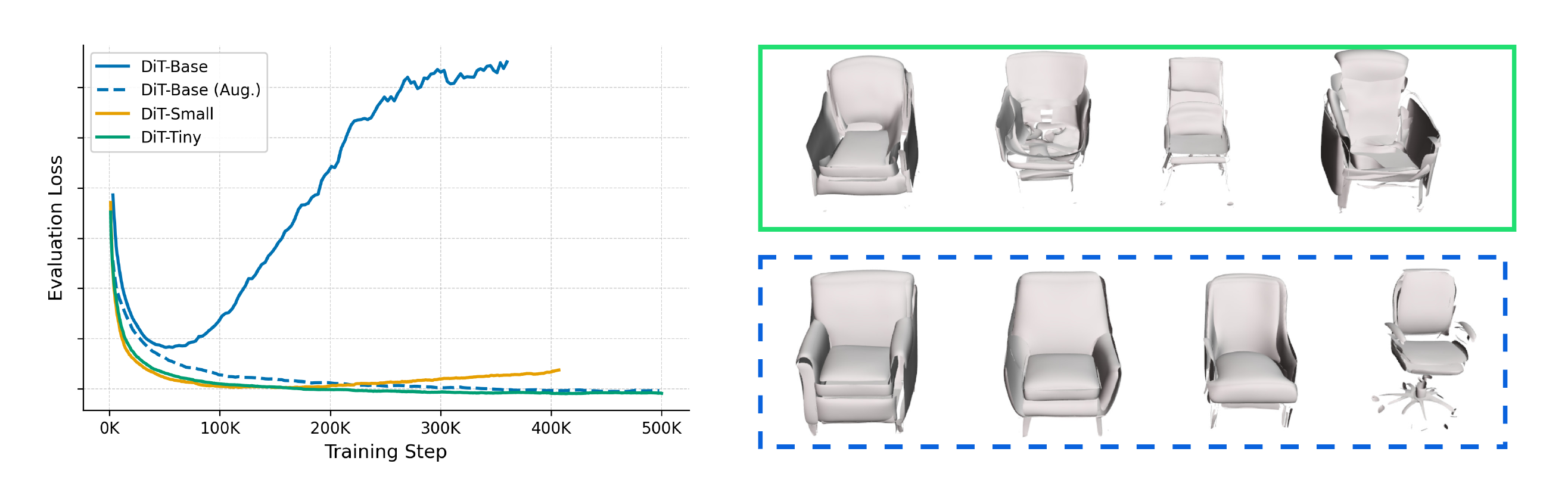}
  \caption{Impact of geometric data augmentation on DiT capacity scaling
    (ABO, 4 classes).
    \textbf{Left:} Evaluation loss vs.\ training step. Without augmentation
    (3{,}800 samples) the higher-capacity DiT-Base diverges and only DiT-Tiny
    trains stably, whereas with augmentation (500K samples) DiT-Base converges
    reliably.
    \textbf{Right:} Qualitative DiT-Base generations, showing that
    augmentation yields better sample quality than the non-augmented model.}
  \label{fig:augmentation_curve}
\end{figure}

With only the original 3{,}800 training samples, only DiT-Tiny converges
stably; DiT-Base shows unstable validation loss (\cref{fig:augmentation_curve}).
Applying our augmentation expands the training set to 500K samples, after
which DiT-Base converges reliably and yields better generation quality than
DiT-Tiny. This demonstrates that simply shrinking the model to achieve
stability sacrifices final quality, and that augmentation is essential for
training larger-capacity models on limited data.

Unlike conventional image augmentations, our approach leverages the structural
constraints of UV-aligned geometry images to generate semantically consistent
variations (3{,}800 $\rightarrow$ 500K), simultaneously improving stability
and generalisation without changing 3D semantics.

\section{Extended Discussion on Saddle-Point Convention in DMS}
\label{sec:supp_saddle}

In Sec.~3.2 of the main paper, we noted that Cases~6 and~9 of Marching
Squares correspond to topologically ambiguous saddle-point configurations,
and that we deterministically treat the two patches as separate independent
regions. Here we provide additional justification.

During UV packing (Sec.~3.1), inter-patch padding ensures that distinct
surface charts are spatially separated on the 2D grid, making it unlikely for
two independent patches to share a single cell in an ambiguous diagonal
configuration. In principle, narrow bridge-like structures within a single
cell could be erroneously split, but such configurations rarely arise at our
operating resolution of $256{\times}256$ with sufficient inter-chart padding.

Importantly, while the topological choice itself is discrete, the coordinates
of each boundary vertex remain continuous functions of the TSDF values
(Eq.~(1) of the main paper), so the backpropagation path is fully preserved
even in ambiguous cases.

\section{Additional Generation Results}
\label{sec:additional}

We present additional qualitative results of our conditional generation
models.

\begin{figure}[tb]
  \centering
  \includegraphics[width=\textwidth]{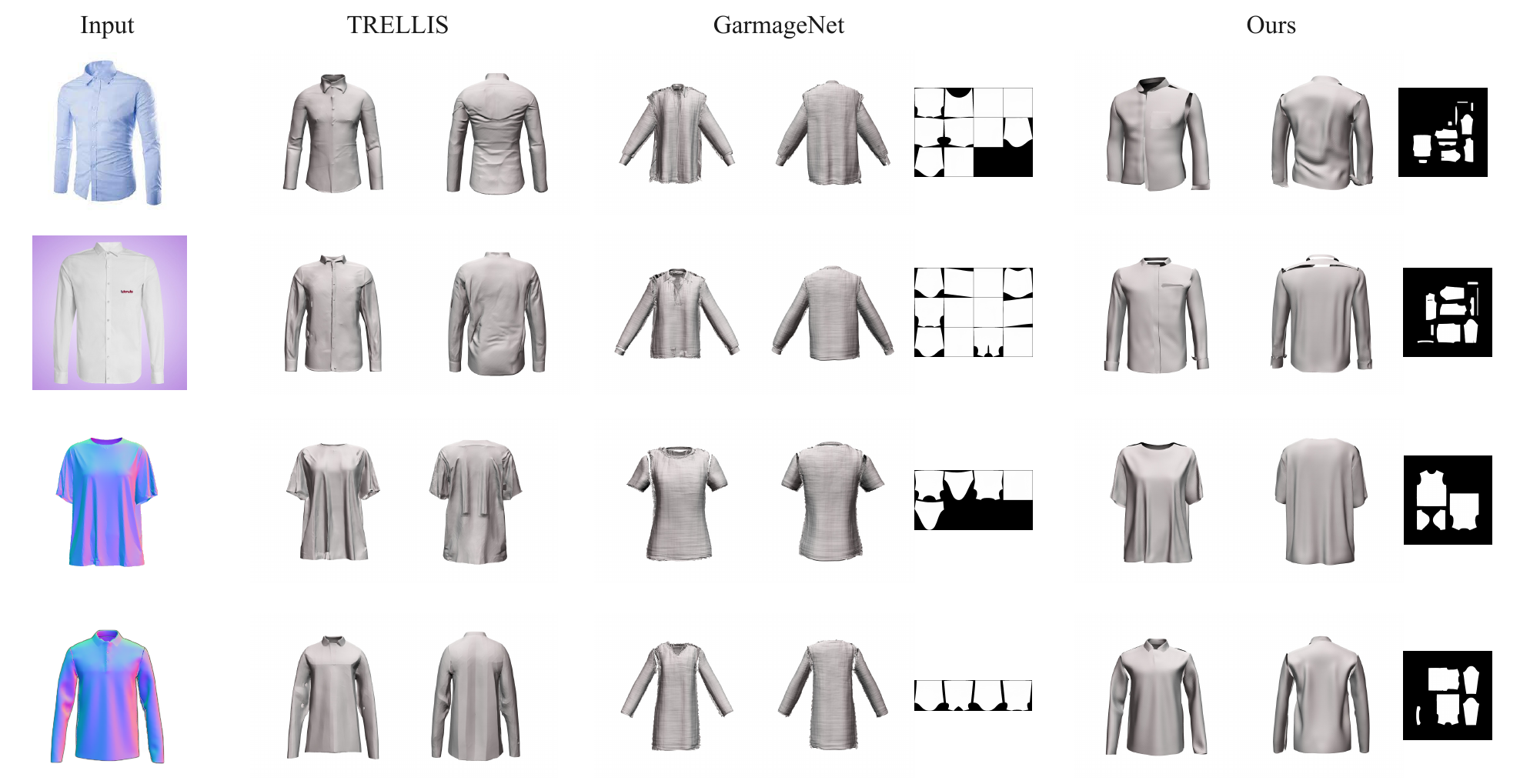}
  \caption{Additional image-conditioned results on GarmageSet (left to right:
    input, TRELLIS, GarmageNet, ours). Our method produces cleaner boundaries
    and thinner shells with significantly fewer vertices (${\sim}23$K
    vs.\ $109$--$526$K).}
  \label{fig:additional_results}
\end{figure}

\paragraph{Image-conditioned generation on GarmageSet.}
\cref{fig:additional_results} compares TRELLIS, GarmageNet, and our method on
GarmageSet samples not shown in the main paper. TRELLIS produces overly thick
meshes due to its watertight assumption; GarmageNet exhibits boundary aliasing
from binary occupancy. Our DiffGI consistently reconstructs cleaner boundaries
and finer fabric details (\eg, collars, hems, sleeve openings) with
significantly fewer vertices.

\begin{figure}[tb]
  \centering
  \includegraphics[width=0.82\textwidth]{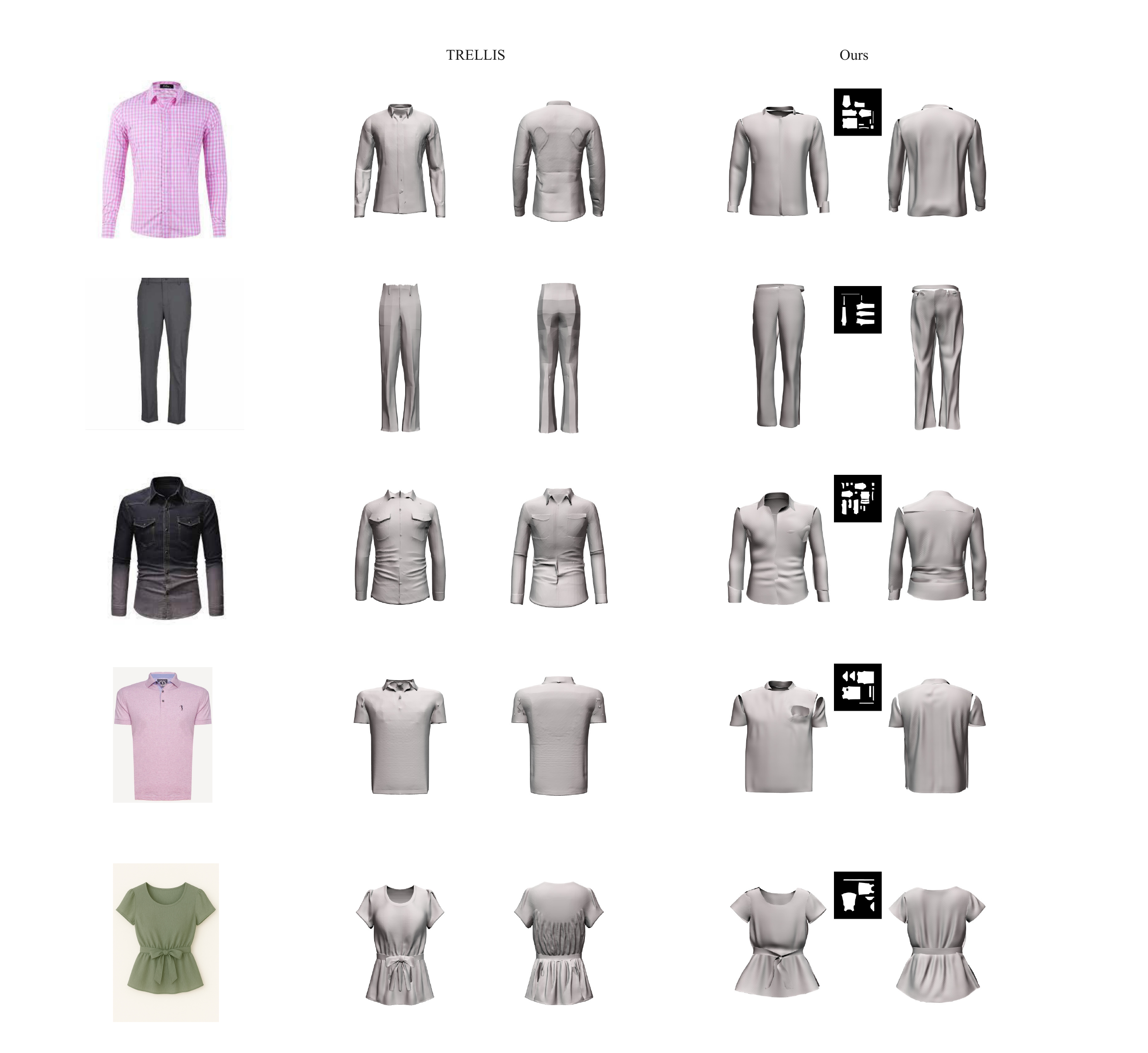}
  \caption{Zero-shot image-to-3D on the WARDROBE
    dataset~\cite{freshersstaff} (not used in training). Our DiffGI better
    matches input silhouettes and preserves open boundaries compared to
    TRELLIS.}
  \label{fig:additional_zeroshot}
\end{figure}

\paragraph{Zero-shot image-to-3D on WARDROBE.}
\cref{fig:additional_zeroshot} shows zero-shot results on real-world clothing
images from the WARDROBE dataset~\cite{freshersstaff}, not used during
training. Our method faithfully recovers thin-shell structures and open
boundaries for unseen categories, demonstrating effective transfer of the
geometric priors learned by DiffGI-VAE.

\begin{figure}[tb]
  \centering
  \includegraphics[width=0.82\textwidth]{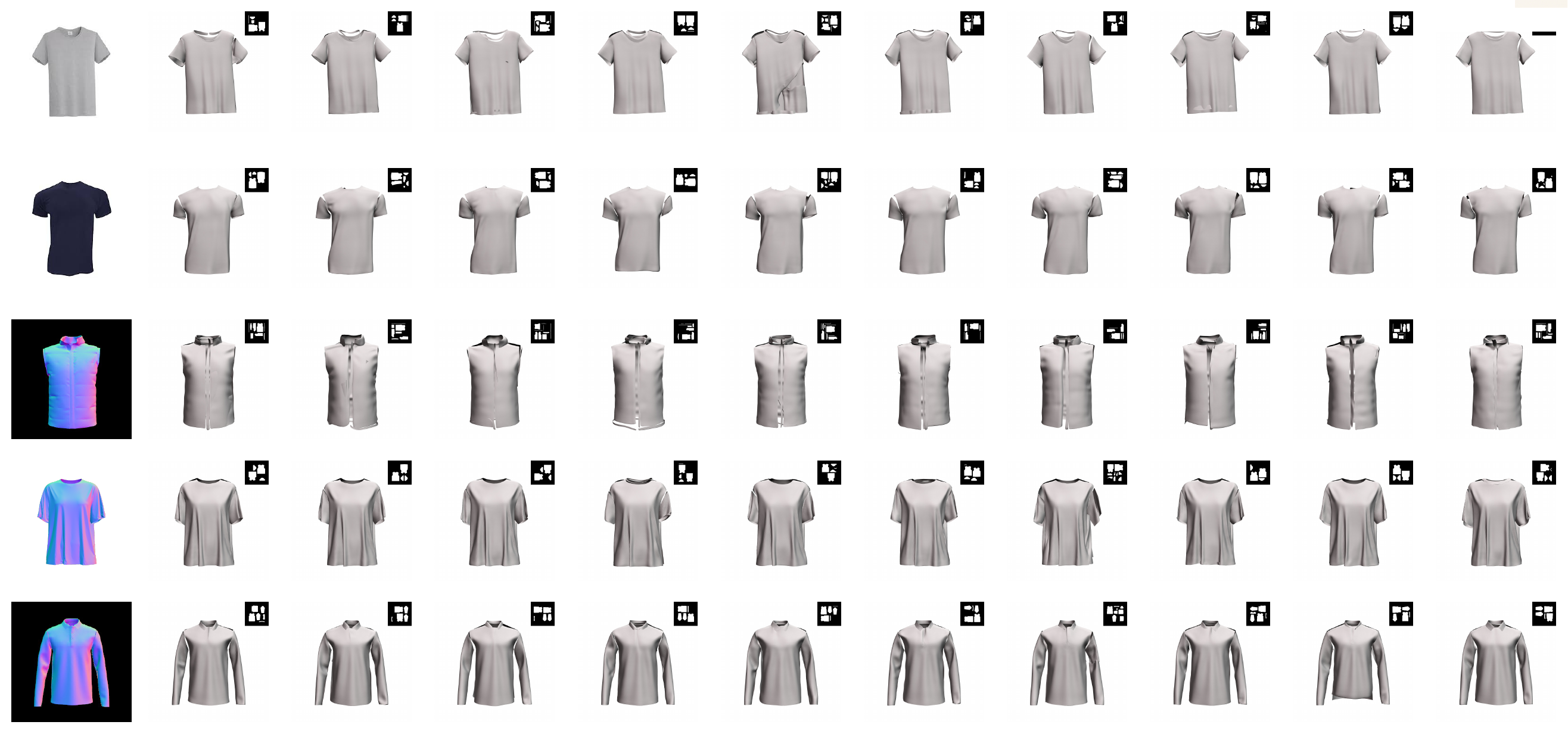}
  \caption{Generation diversity under a fixed condition. Each row shows an
    input image (left) and multiple 3D garments sampled from the same
    condition; insets show the generated geometry-image chart layout of each
    sample. Samples stay consistent with the input while varying naturally
    in chart layout, draping, and wrinkles.}
  \label{fig:additional_multiple}
\end{figure}

\paragraph{Generation diversity under a fixed condition.}
\cref{fig:additional_multiple} shows multiple 3D garments sampled from the
image-conditioned model given the same input image. Across samples, the
generated geometry images adopt different UV chart layouts (insets), yet
every sample decodes into a garment consistent with the input, with natural
variations in draping and wrinkles. This indicates that DiffGI learns a
diverse conditional distribution over both chart layouts and surface
geometry, rather than collapsing to a single mode.

\section{Clarifications on Evaluation Protocol}
\label{sec:supp_eval}

\paragraph{Mesh reconstruction for baselines.}
Each method in Table~1 of the main paper is evaluated with its native mesh
extractor: Differentiable Marching Squares for DiffGI and the Omages-style
tessellation for Omages. GarmageNet's native extraction is a slow multi-stage
non-learning procedure (about 5.1\,s per mesh), so we report it under the same
Omages-style tessellation and additionally under its native official
extraction; DiffGI remains superior under both.

\paragraph{Boundary Chamfer Distance for TRELLIS.}
BCD is reported as N/A for TRELLIS in Table~5 of the main paper because
TRELLIS produces inherently \emph{watertight} meshes without open boundaries,
making BCD inapplicable by definition.

\section{Role of Differentiable Surface Extraction in Enabling
  $\mathcal{L}_{\text{Normal}}$}
\label{sec:supp_normal_loss}

$\mathcal{L}_{\text{Normal}}$ can propagate gradients to the VAE encoder and
decoder only when surface extraction is differentiable; otherwise mesh
reconstruction is a non-differentiable post-processing step that blocks any
3D-space supervision from reaching the network weights. For the TSDF
representation this differentiable extraction is provided by the DMS module,
and for the occupancy variants in the ablation by a differentiable formulation
of the Omages-style tessellation.

This explains the synergy in the ablation (Table~2 of the main paper):
switching from Occ.\ to TSDF \emph{without} $\mathcal{L}_{\text{Normal}}$
already reduces CD by more than half
($1.503 \!\times\! 10^{-3} \!\to\! 0.595 \!\times\! 10^{-3}$), confirming
that continuous TSDF provides the most fundamental gain. Adding
$\mathcal{L}_{\text{Normal}}$ further boosts NC from $0.921$ to $0.961$ and
lowers CD to $0.461 \!\times\! 10^{-3}$, a complementary benefit enabled by
keeping surface extraction differentiable end to end.

\section{Additional Implementation Details}
\label{sec:supp_details}

\paragraph{TSDF map computation.}
The TSDF map is computed entirely in the 2D UV plane and does not use the
3-channel position map. We rasterize the packed UV charts into a binary mask
at $1024{\times}1024$ and apply a 2D Euclidean distance transform, assigning
each pixel its signed distance, in pixels, to the nearest chart contour
(positive inside a chart, negative outside). Because the distance is measured
within the UV plane rather than from a 3D point to the surface, pixels in the
empty background outside all charts, which have no associated 3D position,
still receive a well-defined negative value. This signed distance is clamped
at \textbf{15~pixels} (Mesh-to-DiffGI conversion, Sec.~3.1, Step~4) before the
map is downsampled.

\paragraph{WARDROBE dataset.}
The WARDROBE dataset~\cite{freshersstaff} is used only for qualitative
zero-shot evaluation (\cref{fig:additional_zeroshot}) and is not part of the
quantitative benchmarks (ABO, GarmageSet).

\paragraph{Geometric data augmentation.}
A detailed analysis is provided in \cref{sec:augmentation}.

\paragraph{Diffusion model architectures.}
Architecture specifications for the three conditional generation models
(Sec.~3.4 of the main paper) are summarised in \cref{tab:arch_details}.

\begin{center}
  \begin{minipage}{\textwidth}
  \captionof{table}{Architecture details for the three conditional
    generation models.}
  \label{tab:arch_details}
  \centering
  \footnotesize
  \setlength{\tabcolsep}{4pt}
  \begin{tabular}{@{}lccc@{}}
    \toprule
    & Label-Cond. & Image-Cond. & Occ.-Cond. \\
    \midrule
    Backbone         & DiT-B/2     & DiT-L/2        & UNet-Tiny \\
    Condition        & Class label & DINOv2-Large   & 2D occupancy map \\
    Cond.\ injection & AdaLN-Zero  & Cross-attention & Skip connection \\
    Scheduler        & Flow matching & Flow matching & Flow matching \\
    Latent size      & $32{\times}32{\times}4$ & $32{\times}32{\times}4$
                     & $32{\times}32{\times}4$ \\
    \bottomrule
  \end{tabular}
  \end{minipage}
\end{center}

\end{document}